# A Bio-inspired Modular System for Humanoid Posture Control

Vittorio Lippi, Thomas Mergner, Georg Hettich

*Abstract*—Bio-inspired sensorimotor control systems may be appealing to roboticists who try to solve problems of multi-DOF humanoids and human-robot interactions. This paper presents a simple posture control concept from neuroscience, called disturbance estimation and compensation, DEC concept [1]. It provides human-like mechanical compliance due to low loop gain, tolerance of time delays, and automatic adjustment to changes in external disturbance scenarios. Its outstanding feature is that it uses feedback of multisensory disturbance estimates rather than 'raw' sensory signals for disturbance compensation. After proof-of-principle tests in 1 and 2 DOF posture control robots, we present here a generalized DEC control module for multi-DOF robots. In the control layout, one DEC module controls one DOF (modular control architecture). Modules of neighboring joints are synergistically inter-connected using vestibular information in combination with joint angle and torque signals. These sensory interconnections allow each module to control the kinematics of the more distal links as if they were a single link. This modular design makes the complexity of the robot control scale linearly with the DOFs and error robustness high compared to monolithic control architectures. The presented concept uses Matlab/Simulink (The MathWorks, Natick, USA) for both, model simulation and robot control and will be available as open library.

## I. INTRODUCTION

Postural adjustments (PAs) allow humans to make their voluntary movements smooth and skillful. The adjustments (1) provide the movement buttress that the action-reaction law of physics prescribes, (2) maintain body equilibrium by balancing the body's center of mass (body COM) over the base of support, and (3) cope with interlink coupling torque disturbances from link acceleration (also due to the action-reaction law). The adjustments require coordination across (1)-(3) and between these and the voluntary movements. They participate in the movement and muscle synergies and sensorimotor 'building blocks' [2-5] that help to simplify the complexity given by the high redundancy in the motor system. PA impairment by damage of the cerebellum or sensory systems tends to produce a severely disabling syndrome called ataxia (jerky and dysmetric movements, postural instability; [6]).

There has been recent progress in understanding the neural mechanisms of human postural control [7-10]. This owes to the use of engineering methods that allow relating measured postural responses to exactly known external disturbances in model-based approaches. These models mostly considered human balancing in the sagittal plane, which predominantly occurs around an axis through the ankle joints, and described its biomechanics as that of a single inverted pendulum, SIP.

Among these models, the DEC (disturbance estimation and compensation) model [1, 10] is unique in that it uses sensory-derived internal reconstruction of the external disturbances having impact on the body posture. Model simulation data for various disturbance scenarios and changes in sensor availability were in good agreement with human data [10-14]. The model was re-embodied into a SIP postural control robot [15] and the robot was successfully tested in the human test bed [14].

Further development of the DEC model comprised its extension to double inverted pendulum (DIP) biomechanics with hip and ankle joints and an investigation of the neural control underlying the coordination between these two joints [16]. This work involved a double inverted pendulum, DIP postural control robot. Furthermore, feasibility tests with a 4 DOF agent involving Matlab's SimMechanics toolbox were successful. This led to the here presented generalized DEC module for the control of multi-DOF robots with a modular control architecture. Using one DEC module for each DOF, control complexity linearly increases with the number of DOF.

The next section gives an overview of the DEC control principles, followed by a description of the implementation in a SIP and the generalized modular control of a multi-DOF DEC system. Finally, the DEC library is briefly described and demonstrated by presenting an application. In Conclusion, outstanding results are emphasized and future improvements are outlined.

## II. SYSTEM OVERVIEW

### A. The DEC concept

Figure 1 shows a simplified scheme of the DEC module as it was developed for the SIP control. The module controls joint position of a moving link with respect to a supporting link and consists of three parts:

(A) *Proprioceptive negative feedback loop of joint angular position* (box 'Prop.'). A PID (proportional, integral, derivative) controller provides the torque command (P ≈ $m \cdot g \cdot h$; $m$, body mass; $h$, center of mass= COM height; $g$, gravitational acceleration). The neural time delay of this loop amounts to ≈60 ms (ankle joint).

(B) *Intrinsic stiffness and damping loop of musculoskeletal system* ('passive stiffness' in box '*Biomech*'; it amounts to ≈15% of reflexive stiffness and damping of (A) and

Manuscript received September 6, 2013. Supported by the European
All authors are with the Neurology Department, Freiburg University Clinics, 79106 Freiburg, Germany.
V. L. Author is corresponding author (phone: +49-0761-27052280; fax: +49-0761-27053100; e-mail: Vittorio.lippi@uniklinik-freiburg.de).
T. M. Author (e-mail: mergner@uni-freiburg.de).
G. H. Author (e-mail: georg.hettich@uniklinik-freiburg.de).

feedback gain of (A) and (B) together is unity). Time delays are virtually zero. (A) and (B) together form a servo that, given appropriate control parameters, actuates the joint such that actual joint position equals the desired position (input is displacement trajectory via the Set Point Signal).

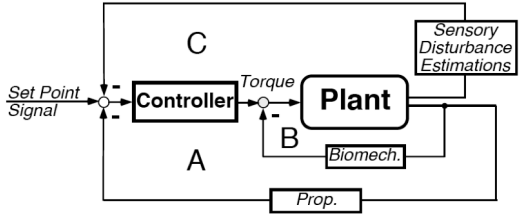

Fig. 1. Simplified scheme of DEC module.

(C) *Disturbance estimation and compensation (DEC) loop*. The DEC loops (four in a complete scheme, see below) estimate the external disturbances through sensor fusions and command via negative feedback the servo (A, B) to produce the joint torques that compensate for the disturbances. Assuming ideal compensation, the servo can function as if there were no disturbances. Note furthermore that no feed forward of plant dynamics is used for the servo (e.g. through an inverse of plant dynamics). Sensory information (mainly vestibular) from the DEC loops upgrades the servo from joint coordinates to space coordinates. Identified lumped human time delay across all three loops (A-C) measured at the controller for the ankle joint amounts to ≈180ms, with the largest share from (C) [9,10].

In the SIP used for the DEC control, the joint (ankle joint) connects the moving link (body) with the supporting link (foot). Postural movements tend to be rather slow, such that centrifugal and Coriolis forces can be neglected. Postural stability is achieved by the DEC feedback for any desired possible joint position, allowing the superposition of voluntary movements with the compensation of external disturbances. The many external events that may have a mechanical impact on body stability are decomposed in, and estimated as external disturbances. Underling these estimations are sensory mechanisms.

Humans use multisensory integration of vestibular signals, vision, touch, and joint proprioception (angle, angular velocity, and force/torque) for their postural control [17]. Studies on human self-motion perception [18] and animal work on sensory processing [19] showed that the central nervous system internally processes physical variables that are not directly available from the sensory organs, but result from sensor fusions.

For example, humans distinguish in the absence of external spatial orientation cues (visual, auditory, and haptic) between body rotation (velocity and position), orientation with respect to the earth vertical, and linear acceleration. They do so by combining input from peripheral vestibular receptor organs (canals, otoliths; see [14]). Furthermore, they may use estimates of the variables for controlling body segments that are distant from the sensor organ in the body. For example, humans may perceive trunk-in-space motion by combining a vestibular head-in-space motion signals with a proprioceptive trunk-to-head (neck) motion signal [18]. Transfer of the space reference also may apply to the other vestibular signals and may be applied to other body segments and even extended to external items that are in firm haptic contact with the body (e.g. the support surface when standing).

The external disturbances and their estimates can be considered from two viewpoints. First, they reflect outside world events that tend to affect the joint torque in certain conditions (e.g. while standing). These events occur in world coordinates and usually in a context dependent way (e.g. ride on "this especially fast escalator"). Corresponding predictions of these estimates may later be re-called from memory and fed forward to the estimation mechanisms where they are fused with the sensory-derived estimates. According to the DEC concept, also self-produced disturbances, such as the gravity effect during voluntary body lean, entail fusions of predicted and sensory derived estimates [1]. The second aspect is that the disturbances affect body stability via the joint torque they produce. The corresponding torque components are referred to as *disturbance torques*. Both aspects will be considered in the next section (III).

The disturbance torques in the SIP scenario are part of the ankle torque

$$T_A = J \cdot \frac{d^2 \alpha_{BS}(t)}{dt^2}, \quad (1)$$

where $J$ represents the body's moment of inertia about the ankle joint (not including the feet) and $\alpha_{BS}$ the body-space angle (primary position: COM projection on ankle joint be vertical, $\alpha_{BS} = 0°$). In the absence of any disturbance, $T_A$ equals the actively produced muscle torque, $T_a$. The disturbance torques add to $T_a$ in the form

$$T_A = (T_g + T_{in} + T_{ext} + T_p) + T_a, \quad (2)$$

where $T_g$ is the gravitational torque, $T_{in}$ the inertial torque, $T_{ext}$ the external torque, and $T_p$ the passive joint torque. $T_g$, $T_{in}$, $T_{ext}$, and $T_p$ challenge the control of $T_A$ (exerted by $T_a$) and are compensated for by $T_a$ [1]. While $T_p$ represents an intrinsic musculoskeletal property, $T_g$, $T_{in}$, and $T_{ext}$ are produced by the neural feedback.

The following section explains the DEC loops in two steps. First, an explanation is given for the simple case of the SIP balancing about the ankle joints. Then, the generalized form for modular control of multi-DOF systems will be presented. Tables I and II give our designations of the DEC module's inputs and outputs, respectively.

## III. DISTURBANCE ESTIMATIONS

Four physical *external* disturbances need to be taken into account for posture control:
(i) Support surface rotation (platform tilt)
(ii) Gravity and other field forces
(iii) Support surface translation (external acceleration)
(iv) Contact forces (external torque)

**(I) DEC OF SUPPORT SURFACE ROTATION**

*SIP scenario.* The support surface tilt produces the foot-space excursion $\alpha_{FS}$ (primary position, level; $\alpha_{FS}= 0°$). Body inertia tends to maintain the primary body-space orientation, which is upright (primary body-foot angle $\alpha_{BF} \approx \alpha_{BS} = 0°$). On the other hand, the servo tends to take the body with the platform in relation to $\alpha_{BF}$. This applies to the passive ankle torque $T_p$, in the form

$$T_p = -K_{P'} \cdot \alpha_{BF} - K_{D'} \cdot \frac{d\alpha_{BF}(t)}{dt} \quad , \quad (3)$$

with $K_{P'}$ representing the passive stiffness (proportional) factor and $K_{D'}$ the passive damping factor. The reflexive part of the servo, loop (A), is commanded by the estimate of $\alpha_{FS}$, $\hat{\alpha}_{FS}$, to maintain the body orientation upright.

To this end, $\hat{\alpha}_{FS}$ combines vestibular and proprioceptive information by a *down channeling* of the vestibular derived space reference from the body to the feet. According to [10], humans achieve this by using the derivatives of $\alpha_{BS}$ and $\alpha_{BF}$ in the form

$$\dot{\alpha}_{FS} = \dot{\alpha}_{BS} - \dot{\alpha}_{BF} \quad . \quad (4)$$

According to [10], there is a subsequent processing of the estimate by a velocity threshold (0.18°/s; with level support, $\dot{\alpha}_{FS}$ tends to be subthreshold, and its noise, mainly from the $\dot{\alpha}_{BS}$ signal, is prevented from entering the control [14]), a scaling factor (G=0.75), and a mathematical integration.

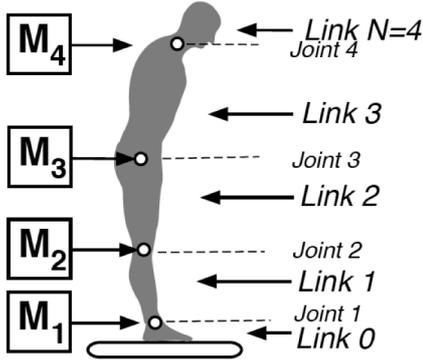

Fig. 2. Conventions used for labeling links, joints and control modules (M).

Implemented in the SIP control, $\hat{\alpha}_{FS}$ 'upgrades' the control from joint to space coordinates.

*Generalized case.* For the n$^{th}$ link in a multi-DOF system (Fig. 2), the link orientation in space is given by $\alpha_n^{SPACE}$. This information is obtained from vestibular input that is down-channeled analogous to (4) through the fusion of vestibular and proprioceptive signals by

$$\alpha_n^{SPACE} = \alpha_{n+1}^{SPACE} - \alpha_{n-1}^{JOINT} \quad . \quad (5)$$

The down-channeling proceeds from the upper most segment $\alpha_{HEAD}^{SPACE}$ that contains the vestibular organs. The tilt of the lowest link in the system (most often the foot), which provides the support for the upper links, is given by

$$\alpha_0^{SPACE} = \alpha_{HEAD}^{SPACE} - \sum_{k=1}^{N} \alpha_k^{JOINT} \quad . \quad (6)$$

Recent evidence from humans [20] suggests that, while the down-channeling to the supporting link occurs through velocity signals as in (4), the further processing in terms of thresholding (dead band discontinuity), gain scaling and integration is performed for the lowest link $\alpha_0^{SPACE}$, from where the tilt estimate is then up channeled for controlling the tilt responses of the upper links. The position of the common *COM* of all the links above the respective joint is calculated (see below). This accounts for the fact that the location of the COM above the joint may change when the configuration of the links changes. The control of each joint can now be viewed as if dealing with a SIP.

**(II) DEC OF GRAVITY DISTURBANCE**

*SIP scenario.* Body lean evokes the gravitational torque $T_g$ that is related to $\alpha_{BS}$ by

$$T_g = m \cdot g \cdot h \cdot \sin(\alpha_{BS}) \quad . \quad (7).$$

The estimate $\hat{T}_g$ uses a vestibular derived $\alpha_{BS}$ signal and includes a detection threshold and gain scaling. For use of $\hat{T}_g$ in the DEC feedback loop, small angle approximation reduces (7) to $T_g=m \cdot g \cdot h \cdot \alpha_{BS}$. Furthermore, $\hat{T}_g$ is divided by $m \cdot g \cdot h$ to obtain an angle equivalent of the torque. As an alternative to use $\hat{T}_g$ one may use directly $\alpha_{BS}$ (in the form of an estimate $\hat{\alpha}_{BS}$).

Note that during support surface tilt with compensation of $\hat{\alpha}_{BS}$ (or $\hat{T}_g$) alone, the body is tilted with the platform; it is the compensation of $\hat{\alpha}_{FS}$ that maintains the body upright.

*Generalized case.* In a DIP or multi-DOF body, $\tau_g$ is calculated by

$$\tau_g = m_n^{UP} g \, CoM_{nx} \quad , \quad (8)$$

where $CoM_{nx}$ is the horizontal component of the position of the center of mass $CoM_n$ of all the segments above the controlled joint. $CoM_n$ is computed performing the weighted average

$$CoM_n = [CoM_{n+1} + L_n \cos(\alpha_n^{SPACE})] m_{n-1}^{UP} \\ + m_n h_n \cos(\alpha_n^{SPACE}) \quad , \quad (9)$$

where $L_n$ is the length of the link controlled by the joint and $h_n$ is the distance of the COM of the n$^{th}$ link from the n$^{th}$ joint.

Analogous to (I), the gravity compensation in each joint comprises all links above this joint, as if dealing with a SIP.

**(III) DEC OF SUPPORT LINEAR ACCELERATION**

*SIP scenario.* Human perception of support surface acceleration may involve various sensory systems and may include sensing of shear forces in the foot soles. However, the corresponding biological knowledge base is still limited. Reference [1] used vestibular information to estimate support surface acceleration, having in mind that vestibular-loss subjects have major problems during such stimulus conditions in the absence of external orientation cues. Support surface acceleration evokes the disturbance torque $T_{in}$ in the form

$$T_{in}' = - \hat{a}_{FS} \cdot m \cdot h \cdot \cos(\alpha_{BS}) \quad , \quad (10)$$

where $\hat{a}_{FS}$ is the estimate of support surface acceleration. $\hat{a}_{FS}$ can be computed from the difference between two vestibular signals, the one of head linear acceleration and the one of the head acceleration due to body rotation (also derived from vestibular input; see [1]. The processing for $\hat{a}_{FS}$ also comprises thresholding and gain scaling.

*Generalized case.* In the case of a multi-DOF system, the external acceleration is computed for each joint. Analogous to the SIP scenario, the part of the vestibular head acceleration signal that is not explained by trunk rotation at the hip or at any joint below is taken to stem from support surface acceleration. This is expressed as

$$a^{EXTERNAL} = a^{VESTIBULAR} - a_n^{SELF} \quad , \quad (11)$$

where the acceleration produced by joint movements is:

$$a_n^{SELF} = a_{n+1}^{SELF} + L_n \frac{d^2}{dt^2}\begin{bmatrix} \sin(\alpha_n^{SPACE}) \\ \cos(\alpha_n^{SPACE}) \end{bmatrix} \quad . \quad (12)$$

The disturbance torque then results from

$$\tau_{acc} = a_x^{EXTERNAL} CoM_{ny} m_n^{UP} + a_y^{EXTERNAL} CoM_{nx} m_n^{UP}. \quad (13)$$

TABLE I.
MODULE INPUTS

| Symbol | Description | Source |
|---|---|---|
| Control signal | Desired $\alpha_n^{SPACE}$, $Com_n$ or $\alpha_n^{JOINT}$ (3 options) | Desired position |
| $CoM_{n+1}$ | Center of mass of the robot over the n+1$^{th}$ joint | n+1$^{th}$ module |
| $m_{n+1}^{UP}$ | Mass of the robot from head to the n+1$^{th}$ joint | n+1$^{th}$ module |
| $\alpha_{n-1}^{SPACE}$ | Up-channeled $\alpha_{n-1}^{SPACE}$ | n-1$^{th}$ module |
| $\alpha_n^{SPACE-DOWN}$ | Down-channeled $\alpha_n^{SPACE}$ | n+1$^{th}$ module |
| $J_{n+1}^{UP}$ | Moment of inertia of the robot from head to the n$^{th}$ joint | n+1$^{th}$ module |
| $a_{n+1}^{ang}$ | Head angular acceleration with respect to the joint n+1$^{th}$ | n+1$^{th}$ module |

**(IV) DEC OF CONTACT FORCE DISTURBANCE**

*SIP scenario.* Led the disturbance torque $T_{ext}$ of equation (2) be the results of a horizontal force $F_{ext}$ exerted on the body by a pull on the clothes at the height $h$, which is above the COM (such that foot-support shear forces may be neglected). $T_{ext}$ is then related to $F_{ext}$ and $T_{BS}$ in the form

$$T_{ext} = F_{ext} \cdot h \cdot \cos(\alpha_{BS}) \quad . \quad (14)$$

An estimate of the external disturbance may be obtained from sensing the amount and location at the body of $F_{ext}$ (and $\alpha_{BS}$). However, having in mind that humans tend to sense centre of pressure (COP) shifts under the feet during such stimuli, studies from our laboratory [1,10,11,13] derived the estimate $\hat{T}_{ext}$ from a sensory measure of COP, as represented in $T_a$, in the form of

$$T_{ext} = T_a - (T_g + T_{in} + T_p - T_A) \quad , \quad (15)$$

accounting for $T_A$, $T_g$, and $T_{in}$ by equations (1), (7), and (10), respectively, and neglecting $T_p$, because it is relatively small. Processing of the $\hat{T}_{ext}$ estimate includes again a detection threshold and, because $T_a$ provides positive feedback, a gain clearly <1 and a low-pass filtering. Humans appear to restrict the use of $\hat{T}_{ext}$ to situations where the contact force stimulus endangers postural stability [13]. Possibly, co-contraction of antagonistic muscles across the involved joints may additionally help $T_{ext}$ compensation as long as the COP shift does not exceed the base of support.

TABLE II
MODULE OUTPUTS

| Symbol | Description | Destination |
|---|---|---|
| $\tau_n$ | Torque produced in the joint n | Joint servo loop |
| $CoM_n$ | Center of mass of the robot over the n$^{th}$ joint | n-1$^{th}$ module |
| $m_n^{UP}$ | Mass of the robot from head to the n$^{th}$ joint | n-1$^{th}$ module |
| $\alpha_n^{SPACE}$ | Up-channeled $\alpha_n^{SPACE}$ | n+1$^{th}$ module |
| $\alpha_n^{SPACE-DOWN}$ | Down-channeled $\alpha_n^{SPACE}$ | n-1$^{th}$ module |
| $J_n^{UP}$ | Moment of inertia of the robot from head to the n$^{th}$ joint | n-1$^{th}$ module |
| $a_n^{ang}$ | Head angular acceleration with respect to the n$^{th}$ joint | n-1$^{th}$ module |

*Generalized case.* For convenience, the subscripts in (15) were changed to superscripts, allowing to denote the number of the module by the subscript. With this modification, (15) takes the form

$$\tau_{ext} = \tau_n^a - \tau_n^g - \tau_n^{in} - \tau_n^p + \tau_n^A \quad . \quad (16)$$

With the term $J_n^{UP}$ representing the moment of inertia of all the segments over the controlled joint, equation (1) takes the form

$$\tau_n^A = \frac{d}{dt}(\dot{\alpha}_n^{SPACE} J_n^{UP}) \quad , \quad (17)$$

which takes into account that also $J_n^{UP}$ may change in the case of a robot with several DOFs.

In order to keep the computation of $J_n^{UP}$ as simple as possible by exchanging only one variable between blocks, the moment of inertia $J_n^{UP*}$ is computed around the axis passing through the center of mass of the whole group of segments from the $n^{th}$ segment to the *head* in the form

$$J_n^{UP*} = (J_{n+1}^{UP*} + m_{n+1}^{UP}||CoM_{n+1} - CoM_n||^2) + \\ + J_n + m_n||CoM_n^{LINK} - CoM_n||^2 \quad , \quad (18)$$

where $CoM_n^{LINK}$ is the center of mass of the $n^{th}$ link equal to

$$CoM_n^{LINK} = l_n \begin{bmatrix} \sin(\alpha_n^{SPACE}) \\ \cos(\alpha_n^{SPACE}) \end{bmatrix} \quad . \quad (19)$$

$J_n^{UP*}$ is then down-channeled to the n-1$^{nt}$ block, while $J_n^{UP}$, used in (17), is computed as

$$J_n^{UP} = J_n^{UP*} + m_n^{UP}||CoM_n||^2 \quad . \quad (20)$$

## IV. SOFTWARE LIBRARY

The software library consists of a Matlab/Simulink block that implements a single DEC module containing the servo and the above described disturbance estimates. An interactive mask allows the user to specify the block position in a multi-link system (as top or bottom/supporting, or intermediate link). The physical features such as link height, link mass, height of link COM, etc. are input into the block as anthropometric parameters. Each block has input and output ports to exchange data with the neighboring blocks above and below (see Tables I and II) and allows modifying the processing of the estimates in terms of thresholding, gain scaling, etc. The PID controller parameters and the passive stiffness and damping parameters of the servo are set after deciding the target variables (COM-joint orientation of the above link, or link orientation with respect to earth vertical, or joint angle).

The software will be made available as an open library in the internet.

## V. CASE STUDY

Simulations of the modular control concept were performed with a four link humanoid agent in Matlab/Simulink. The links were feet (fixed to the ground), shank, thigh and trunk (here HAT; head, arms and torso), interconnected by the ankle joint, the knee joint and the hip joint, respectively. Stimuli were applied in the sagittal plane, which allowed us to use a planar triple inverted pendulum biomechanical model. The humanoid's biomechanical parameters corresponded to human anthropometric measures [21]. The above-described DEC modules were used to control the ankle, knee and hip joints in a modular way. The control parameters were adapted from [20]. Accounting for the fact that humans tend to stiffen the knee joints during our tests (see below), a high level of passive stiffness was used for this joint.

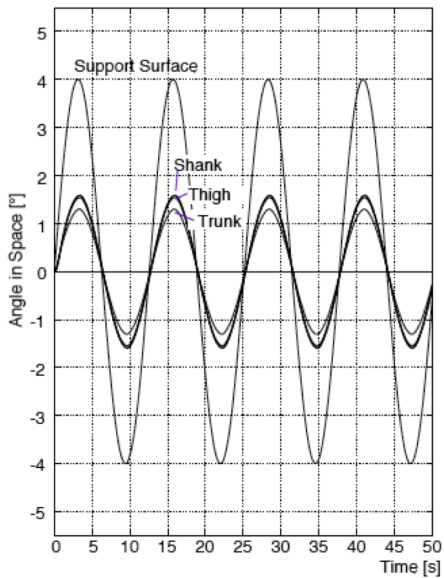

Fig. 3. Responses of the simulated four link humanoid agent to sinusoidal support surface rotation (0.08 Hz). The combined action of the gravitational torque compensation, support tilt compensation and support linear acceleration compensation tends to maintain the body upright, but this only partially due to human-like compensation gains (<1).

Two experiments were performed. In the first experiment, the agent balanced sinusoidal ±4° support surface tilts (Fig. 3). In this test, the agent used the gravitational torque compensation and the support surface tilt compensation for controlling the ankle joint, and these two compensation together with the support surface acceleration compensation for controlling the knee and the hip joints. Compensation gains were set to human-like values (<1). This entailed the under-compensation shown in Fig. 3, with the trunk being compensated slightly better than the two leg segments.

In the second experiment, the agent performed a voluntary forward trunk lean of 4° (Fig. 4). This experiment tested whether the control would produce the human hip-ankle coordination, which consists of a compensatory backward lean of the leg segments such that the body COM is maintained over the ankle joint (see [20]). Additionally, this coordination neutralizes most of the coupling torques exerted by the trunk bending on the leg segments. The voluntary movement is associated with predictions of the disturbance estimates, which have unity gain, and are fused with the sensory-derived disturbance estimates (see [1]). This explains the almost perfect performance in Fig. 4.

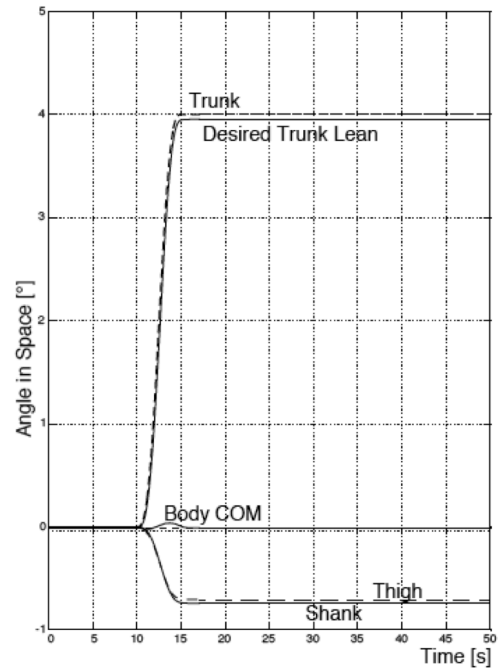

Fig. 4. Behavior of the four link humanoid agent performing a voluntary forward trunk lean in the hip joint. Note that the leg links Thigh and Shank are leaning backwards, so that the Body COM remains above the ankle joint. This ankle-hip coordination automatically arises from the interaction between the hip and ankle control modules and the agent's biomechanics.

## VI. CONCLUSION

Upgrading the DEC concept from using one module in the SIP control to the modular control architecture in a multi-

DOF system is possible, because the DEC concept allows controlling each joint as if it were dealing with a SIP. In particular:

(I) The support surface tilt DEC estimates take the rotation of each supporting link as a tilt disturbance for the upper links. Controlled is the orientation of the links' COM above the supporting joint with respect to earth vertical.

(II) The gravity DEC estimates compensate for the gravitational torque produced by the upper links' total COM.

(III) The support surface linear acceleration DEC estimates compensate for the acceleration effect occurring at the top of any supporting link. The effect is produced as joint torque by the inertial force of all upper links. Noticeably, this compensation also includes up-going coupling forces effects (concerning down-going coupling forces, see [20]).

(IV) The contact force DEC estimates compensate for the evoked torque in the supporting joint, taking into account the moment of inertia of the above (supported) links.

The present concept attributes postural responses to unforeseen external disturbances to sensory mechanisms and feedback. A sensory network of down- and up-going spinal pathways from the brainstem and back to it and to higher CNS centers (e.g. cerebellum) carrying vestibular signals and receiving spinal proprioceptive input has been demonstrated in animal work [22]. Principles of how predictions of disturbance estimates, centrally derived and fed forward during voluntary movements, may be fused with sensor-derived disturbance estimates have been suggested in Reference [1]. Thresholding and gain scaling of estimates have been attributed to sensory noise in Reference [14].

Human movement coordination such as the hip-ankle coordination occurring during voluntary trunk bending (Fig. 4) or balancing of support surface tilt [20] may emerge as automatic 'postural adjustments' from DEC mechanisms. Given moderate disturbances and full foot support, ankle and hip responses fulfill the criteria of the human ankle and hip strategy (see [17]).

In a previous study, we compared the DEC concept with the classical control approach that uses extended observers for disturbance estimation [23]. This solution worked in simulations, but had problems to deal with inaccuracies of sensors and actuation when implemented into the robot. A later solution that included the vestibular system into a standard engineering approach was more successful in terms of stability, but not in terms of human-like responses [24].

Empirically, model simulations and experiments with the robot demonstrated stability of the system [20]. A mathematical generalized demonstration is beyond the scope of the present paper and will be postponed to a specific treatment.

Current work on DEC tries to include visual information into the sensory fusions, to combine sagittal and frontal plane DEC modules, to deal with the four-bar linkage of biped stance (2 legs-ground-pelvis) in the frontal plane, to deal with the shifting of body weight between legs during walking, and to explore further examples of human movement coordination.

REFERENCES

[1] T. Mergner, "A neurological view on reactive human stance control," *Annu Rev Control*, vol. 34, pp. 177–198, 2010.
[2] M. C. Tresch, P. Saltiel, and E. Bizzi, "The construction of movement by the spinal cord," *Nature Neuroscience*, vol. 2, 162–167, 1999.
[3] Y. P. Ivanenko, G. Cappellini, N. Dominici, R. E. Poppele, and F. Lacquaniti, "Coordination of locomotion with voluntary movements in humans," *Journal of Neuroscience*. vol. 25, pp. 7238–7253, 2005.
[4] L. H. Ting, "Dimensional reduction in sensorimotor systems: a framework for understanding muscle coordination of posture," *Progress in Brain Research*, vol. 165, pp. 301–325, 2007.
[5] A. d'Avella and D. K. Pai, "Modularity for sensorimotor control: evidence and a new prediction. *Journal of Motor Behavior*, vol. 42, pp. 361-369, 2010.
[6] A. J. Bastian, "Mechanisms of ataxia," *Physical Therapy*, vol. 77, pp. 672-675, 1997.
[7] R. Johansson and M. Magnusson, "Human postural dynamics," *CRC Crit. Rev. Biomed. Eng.*, vol. 18, pp. 413-437, 1991.
[8] H. van der Kooij, R. Jacobs, B. Koopman, and H. Grootenboer, "A multisensory integration model of human stance control," *Biological Cybernetics*, vol. 80, pp. 299–308, 1999.
[9] R. J. Peterka, "Sensorimotor integration in human postural control," *Journal of Neurophysiology*, vol. 88, pp. 1097–1118, 2002.
[10] C. Maurer, T. Mergner, and R. J. Peterka, "Multisensory control of human upright stance," *Experimental Brain Research*, vol. 171, pp. 231–250, 2006.
[11] T. Mergner, C. Maurer, and R. J. Peterka "A multisensory posture control model of human upright stance," *Progress in Brain Research*, vol. 142, pp. 189-201, 2003.
[12] G. Schweigart and T. Mergner, "Human stance control beyond steady state response and inverted pendulum simplification. *Experimental Brain Research*, vol. 185, pp. 635-653, 2008.
[13] C. Cnyrim, T. Mergner, and C. Maurer, "Potential role of force cues in human stance control," *Experimental Brain Research*, vol. 194, pp. 419–433, 2009.
[14] T. Mergner, G. Schweigart, and L. Fennell, "Vestibular humanoid postural control," *Journal of Physiology - Paris*, vol. 103, pp. 178–194, 2009.
[15] T. Mergner, F. Huethe, C. Maurer, and C. Ament, C, "Human equilibrium control principles implemented into a biped robot," In T. Zielinska and C. Zielinski (eds.) *Robot Design, Dynamics, and Control* (Romansy 16), CISM Courses and Lectures, vol. 487, pp. 271-279, 2006.
[16] G. Hettich, T. Mergner, and L. Fennel, "Double inverted pendulum model of reactive human stance control." In: Multibody Dynamics 2011, ECCOMAS Thematic Conference., 2011. Brussels, Belgium, 4-7 July 2011. (http://www.posturob.uniklinik-freiburg.de).
[17] B. Horak and J.M. Macpherson, "Postural orientation and equilibrium', in L. Rowell and J. Shepherd (ed.) Handbook of Physiology, New York: Oxford University Press, 1996, pp 255-292.
[18] T. Mergner, W. Huber, and W. Becker, "Vestibular-neck interaction and transformations of sensory coordinates," *Journal of Vestibular Research*, vol. 7, pp. 119–135, 1997.
[19] G. Bosco and R.E. Poppele, "Representation of multiple kinematic parameters of the cat hindlimb in spinocerebellar activity," *Journal of Neurophysiology*, vol. 78, pp.1421–1432, 1997.
[20] G. Hettich, L. Assländer, A. Gollhofer, and T. Mergner, "Human hip-ankle coordination emerging from multisensory feedback control," Submitted.
[21] D. A. Winter, *Biomechanics and motor control of human movement. (2nd ed.)*. New York: Wiley, 1990.
[22] J. D. Coulter, T. Mergner, and O. Pompeiano, "Effects of static tilt on cervical spinoreticular tract neurons," *Journal of Neurophysiology*, vol. 39, pp. 45-62, 1976.
[23] K.A. Tahboub and T. Mergner, "Biological and engineering approaches to human postural control," *Journal of Integrated Computer Aided Engineering*, vol. 14, pp. 15-31, 2007.
[24] K.A. Tahboub, "Biologically-inspired humanoid postural control," *Journal of Physiology - Paris*, vol. 103, pp. 195–210, 2009.